%% file: iclr2015.tex
\documentclass{article} 
\usepackage{iclr2015,times}
\usepackage{hyperref}
\usepackage{url}
\usepackage{graphicx}

\title{Learning Longer Memory in \\ Recurrent Neural Networks}

\author{
Tomas Mikolov,~Armand Joulin,~Sumit Chopra,~Michael Mathieu \& Marc'Aurelio Ranzato \\
Facebook  Artificial Intelligence Research\\
770 Broadway\\
New York City, NY, 10003, USA \\
\texttt{\{tmikolov,ajoulin,spchopra,myrhev,ranzato\}@fb.com}
}

\iclrfinalcopy 

\begin{document}

\maketitle

\begin{abstract}
Recurrent neural network is a powerful model that learns
temporal patterns in sequential data. 
For a long time, it was believed that recurrent networks are
difficult to train using simple optimizers, such as
stochastic gradient descent, due to the so-called
vanishing gradient problem.
In this paper, we show that learning longer term patterns in
real data, such as in natural language, is perfectly possible using gradient descent.
This is achieved by using a slight structural
modification of the simple recurrent neural network architecture. 
We encourage some of the hidden units to change their state slowly by
making part of the recurrent weight matrix 
close to identity, thus forming a kind of longer term memory. 
We evaluate our model on language modeling tasks on benchmark 
datasets, where we obtain similar performance to the much more complex 
Long Short Term Memory (LSTM) 
networks~\citep{hochreiter1997long}.
\end{abstract}

\input{intro.tex}

\input{model.tex}

\input{experiments.tex}

\input{conclusion.tex}

\bibliography{iclr2015}
\bibliographystyle{iclr2015}

\end{document}

%% file: intro.tex
\section{Introduction}

Models of sequential data, such as natural language, speech and video, 
are the core of many machine learning applications. 
This has been widely studied in the past with approaches taking their roots 
in a variety of fields~\citep{goodman2001bit, young1997htk, koehn2007moses}.
In particular, models based on neural networks have been very successful recently, 
obtaining state-of-the-art
performances in automatic speech recognition~\citep{dahl2012context}, 
language modeling~\citep{mikolov2012statistical} and 
video classification~\citep{simonyan2014two}.
These models are mostly based on two families of neural networks: 
feedforward neural networks and recurrent neural networks.

Feedforward architectures such as time-delayed neural networks 
usually represent time explicitly with a fixed-length
window of the recent history~\citep{rumelhart1985learning}. 
While this type of models work well in practice, fixing the window size makes 
long-term dependency harder to learn and can only be done at the 
cost of a linear increase of the number of parameters.

The recurrent architectures, on the other hand, represent time recursively.
For example, in the simple recurrent network (SRN)~\citep{elman1990finding}, the state of the hidden layer 
at a given time is conditioned on its previous state.
This recursion allows the model to store complex signals for arbitrarily long time periods, as the
state of the hidden layer can be seen as the memory of the model.
In theory, this architecture could even encode a ``perfect'' memory
by simply copying the state of the hidden layer over time. 

While theoretically powerful, these recurrent models were widely considered to be hard to train
due to the so-called vanishing and exploding gradient problems~\citep{hochreiter1998vanishing,bengio1994learning}. 
\citet{mikolov2012statistical} showed how to avoid the exploding gradient problem
by using simple, yet efficient strategy of gradient clipping. This allowed to efficiently train 
these models on large datasets by using only simple tools such as stochastic gradient
descent and back-propagation through time~\citep{williams1995gradient,werbos1988generalization}. 

Nevertheless, simple recurrent networks still suffer from the vanishing 
gradient problem:
as gradients are propagated back through time,
their magnitude will almost always exponentially shrink.
This makes memory of the SRNs focused only on short term patterns,
practically ignoring longer term dependencies.
There are two reasons why this happens.  First, standard nonlinearities such as the sigmoid function
have a gradient which is close to zero almost everywhere. 
This problem has been partially 
solved in deep networks by using the rectified linear units (ReLu)~\citep{nair2010rectified}.
Second, as the gradient is backpropagated through time, its magnitude
is multiplied over and over by the recurrent matrix. If the eigenvalues
of this matrix are small (i.e., less than one), the gradient will converge to zero
rapidly.
Empirically,  
gradients are usually close to zero after 5 - 10 steps of backpropagation.
This makes it hard for simple recurrent neural networks to learn any long term patterns. 

Many architectures were proposed to deal with the vanishing gradients. 
Among those, the long short term memory (LSTM) recurrent neural network~\citep{hochreiter1997long} is a modified
version of simple recurrent network which has obtained promising results
on hand writing recognition~\citep{graves2009offline} and phoneme classification~\citep{graves2005framewise}. 
LSTM relies on a fairly sophisticated structure made of gates which control flow of information
to hidden neurons. This allows the network to potentially remember information for longer periods.
Another interesting direction which was considered is to exploit the structure of the Hessian matrix with respect to 
the parameters to avoid vanishing gradients.
This can be achieved by using second-order methods designed for non-convex objective functions (see section 7 in \citet{lecun1998efficient}).
Unfortunately, there is no clear theoretical justification why using the Hessian matrix would help, nor there is, to the best of our knowledge, 
any conclusive thorough empirical study on this topic.


In this paper, we propose a simple modification of the SRN to partially solve the vanishing gradient problem. 
In Section \ref{sec:model}, we demonstrate that by simply constraining a part of the recurrent matrix to be close to identity, we can drive
some hidden units, called \emph{context units} to behave like a cache model which can capture long term
information similar to the topic of a text~\citep{kuhn1990cache}. In Section \ref{sec:exp}, we show that our model can obtain competitive performance
compared to the state-of-the-art sequence prediction model, LSTM, on language modeling datasets.


%% file: model.tex
\section{Model}
\label{sec:model}
\subsection{Simple recurrent network}

 \begin{figure}[h!]
 \begin{center}
 \begin{tabular}{cc}
  \includegraphics[width=0.25\textwidth]{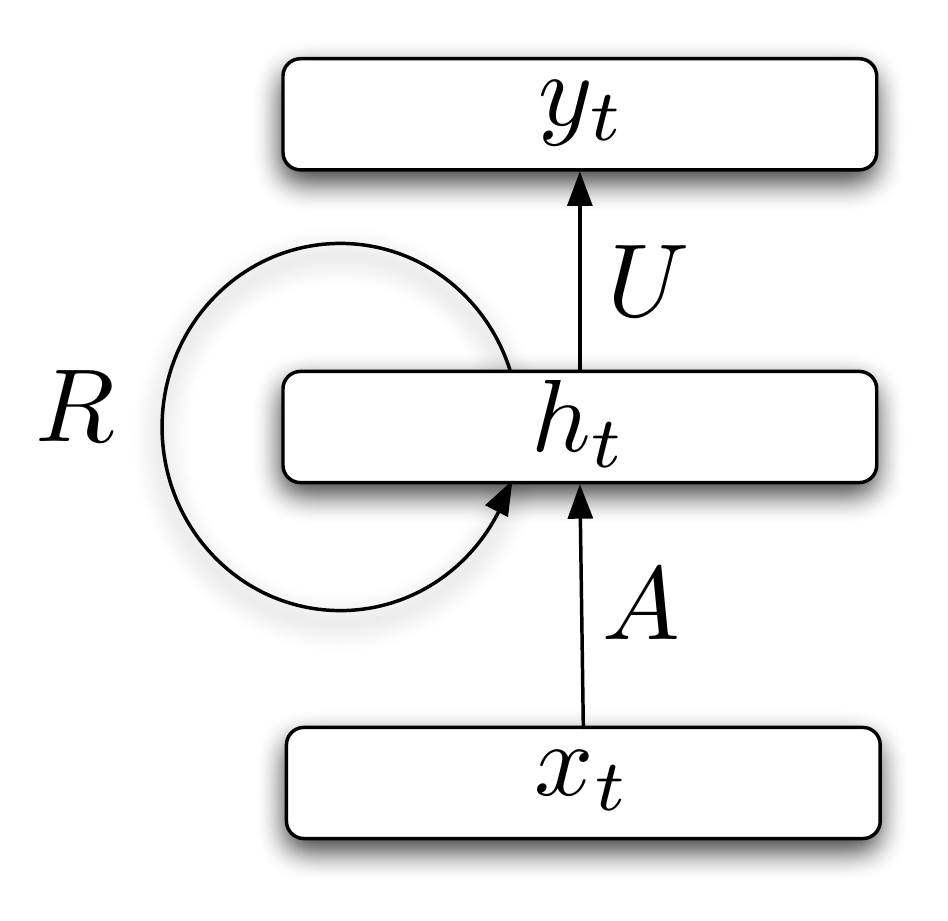}
  &
  \includegraphics[width=0.5\textwidth]{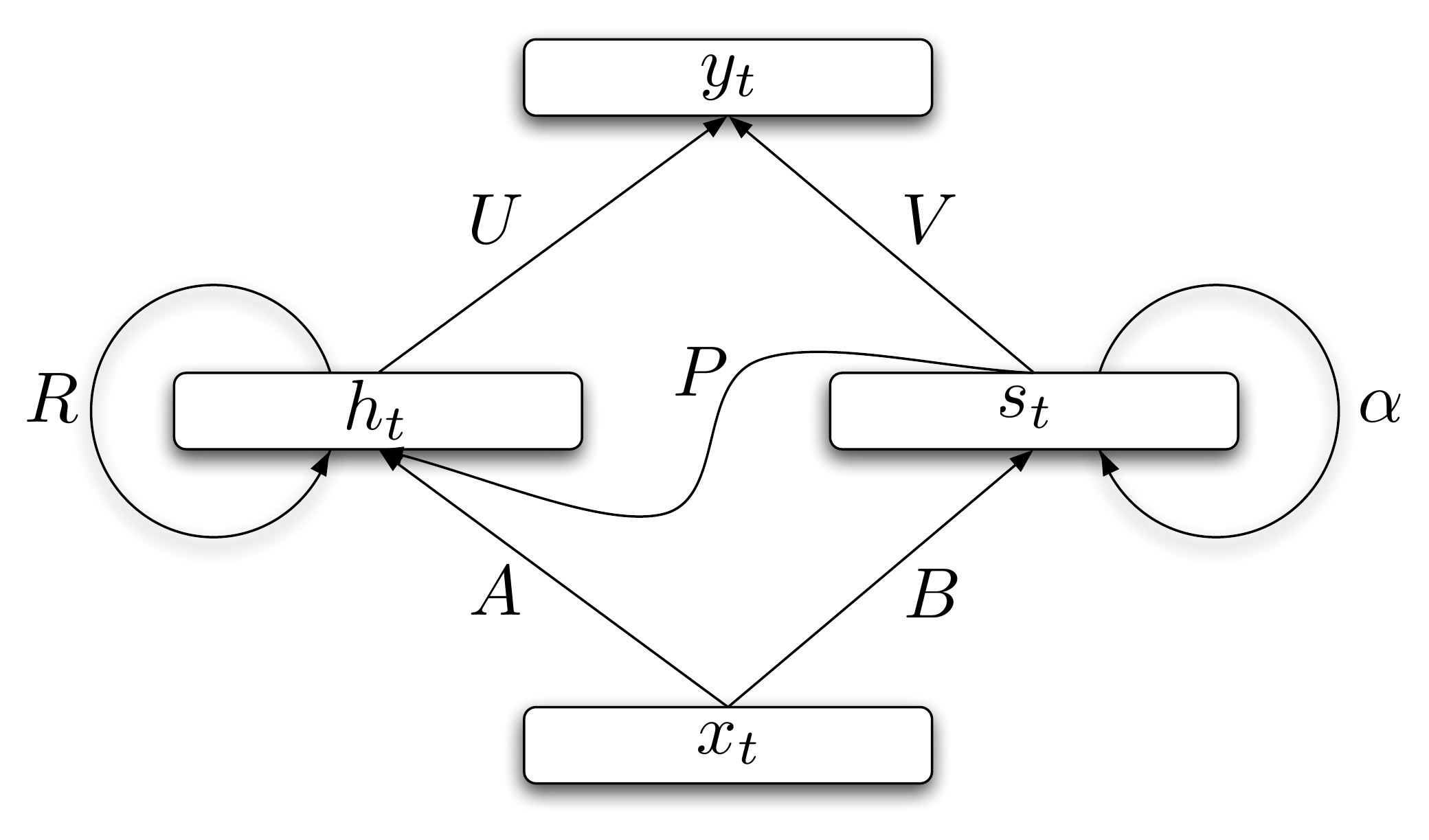}
\\
~~~~~~~~~(a) & (b)
\end{tabular}
\caption{(a) Simple recurrent network. (b) Recurrent network with context features.}
\label{fig:model}
 \end{center}
 \end{figure}

We consider sequential data that comes in the form
of discrete tokens, such as characters or words. We assume a fixed dictionary containing $d$ 
tokens. Our goal is to design a model which is able to predict the next token in the sequence given its past. 
In this section, we describe the simple recurrent network (SRN) model popularized by \citet{elman1990finding}, and which is the cornerstone of this work. 

A SRN consists of an input layer, a hidden layer with a recurrent connection and an output layer (see Figure \ref{fig:model}-a).
The recurrent connection allows the propagation through time of information about the state of the hidden layer. Given a sequence
of tokens, a SRN takes as input the one-hot encoding $x_t$ of the current token and predicts the probability $y_t$ of next one.
Between the current token representation and the prediction, there is a hidden layer with $m$ units which store additional
information about the previous tokens seen in the sequence. More precisely, at each time $t$, 
the state of the hidden layer $h_t$ is updated based on its previous state $h_{t-1}$ and the encoding $x_t$ of the current token,
according to the following equation:

\begin{equation}
h_t = \sigma \left( A x_t + R h_{t-1} \right),\label{eq:h}
\end{equation}

where $\sigma(x) = 1/ (1+\exp(x))$ is the sigmoid function applied coordinate wise,
$A$ is the $d\times m$ token embedding matrix and $R$ is the $m\times m$ matrix of
recurrent weights. Given the state of these hidden units, the network then outputs the
probability vector $y_t$ of the next token, according to the following equation:

\begin{equation}
y_t = f \left(Uh_t \right),
\end{equation}

where $f$ is the soft-max function and $U$ is the $m\times d$ output matrix. In some cases, the size $d$ of the dictionary can be significant (e.g., more
than 100K tokens for standard language modeling tasks) 
and computing the normalization term of the soft-max function is often the bottle-neck of this type of architecture. 
A common trick introduced in \citet{goodman2001classes} is to 
replace the soft-max function by a hierarchical soft-max. 
We use a simple hierarchy with two levels, by binning the tokens into $\sqrt{d}$ clusters 
with same cumulative word frequency~\citep{mikolov2011extensions}. 
This reduces the complexity of computing the soft-max from $O(hd)$ to about $O(h\sqrt{d})$, 
but at the cost of lower performance (around 10$\%$ loss in perplexity). We will mention explicitly when we use this approximation in the experiments. 

The model is trained by using stochastic gradient descent method with back-propagation through time~\citep{rumelhart1985learning, williams1995gradient,werbos1988generalization}. 
We use gradient renormalization to avoid gradient explosion. 
In practice, this strategy is equivalent to gradient clipping since gradient
explosions happen very rarely when reasonable hyper-parameters are used.
The details of the implementation are given in the experiment section.

It is generally believed that using a strong nonlinearity is
necessary to capture complex patterns appearing in real-world data. In particular,
the class of mapping that a neural network can learn between the input space and the
output space depends directly on these nonlinearities (along with the number of hidden
layers and their sizes). However, these nonlinearities also introduce the so-called
vanishing gradient problem in recurrent networks. The vanishing gradient problem states
that as the gradients get propagated back in time, their magnitude quickly shrinks close to
zero. This makes learning longer term patterns difficult, resulting in
models which fail to capture the surrounding context. In the next section,
we propose a simple extension of SRN to circumvent this problem, yielding a
model that can retain information about longer context.

\subsection{Context features}

In this section, we propose an extension of SRN by adding a hidden layer specifically designed
to capture longer term dependencies. We design this layer following two observations:  (1) the
nonlinearity can cause gradients to vanish, (2) a fully connected hidden layer changes its state completely at every time step.

SRN uses a fully connected recurrent matrix which allows complex patterns to be propagated through time but suffers 
from the fact that the state of the hidden units changes rapidly at every time step. 
On the other hand, using a recurrent matrix equal to identity and removing the nonlinearity 
would keep the state of the hidden layer constant, and every change of the state would have to come from
external inputs.
This should allow to retain information for longer period of time.
More precisely, the rule would be:

\begin{equation}
s_t = s_{t-1} + B x_t,
\end{equation}

where $B$ is the $d\times s$ context embedding matrix. 
This solution leads to a model which cannot be trained efficiently. 
Indeed, the gradient of the recurrent matrix would never vanish, which would require propagation of the gradients 
up to the beginning of the training set.

Many variations around this type of memory 
have been studied in the past (see \citet{mozer1993neural} for an overview of existing models).
Most of these models are based on SRN with 
no off-diagonal recurrent connections between the hidden units. They differ in how the diagonal weights of the recurrent matrix
are constrained. 
Recently, \citet{pachitariu2013regularization} showed that this type of architecture can achieve performance similar to a full SRN
when the size of the dataset and of the model are small.
This type of architecture can potentially retain information about longer term statistics, 
such as the topic of a text, but it does not scale well to larger datasets~\citep{pachitariu2013regularization}.
Besides, it can been argued that purely linear SRNs with learned self-recurrent weights will perform very similarly
to a combination of cache models with different rates of information decay~\citep{kuhn1990cache}.
Cache models compute probability of the next token given a bag-of-words (unordered) representation
of longer history. They are well known to perform strongly on small datasets~\citep{goodman2001bit}.
\citet{mikolov2012context} show that using such contextual features as 
additional inputs to the hidden layer leads to a significant improvement in performance over the regular SRN. 
However in their work, the contextual features are pre-trained using standard NLP techniques and not learned
as part of the recurrent model.

In this work, we propose a model which learns the contextual features using stochastic gradient descent.
These features are the state of a hidden layer associated with a diagonal recurrent matrix similar to the one presented in~\citet{mozer1993neural}.
In other words, our model possesses both a fully connected recurrent matrix to produce a set of quickly changing hidden units, and a diagonal
matrix that that encourages the state of the context units to change slowly (see the detailed model in Figure~\ref{fig:model}-b). 
The fast layer (called \emph{hidden layer} in the rest of this paper) can learn representations similar to n-gram models, 
while the slowly changing layer (called \emph{context layer}) can learn topic information, similar to cache models.
More precisely, denoting by $s_t$ the state of the $p$ context units at time $t$, the update rules of the model are:

\begin{eqnarray}
s_t &=& (1- \alpha) B x_t + \alpha s_{t-1},\label{eq:s}\\
h_t &=& \sigma \left( P s_t + A x_t + R h_{t-1} \right),\label{eq:h2}\\
y_t &=& f \left( U h_t + V s_t \right)
\end{eqnarray}

where $\alpha$ is a parameter in $(0,1)$ and $P$ is a $p\times m$ matrix. Note that there is no nonlinearity
applied to the state of the context units. The contextual hidden units can be
seen as an exponentially decaying bag of words representation of the history.
This {\it exponential trace memory} (as denoted by~\citet{mozer1993neural}) has been
already proposed in the context of simple recurrent networks~\citep{jordan1987,mozer1989focused}.

A close idea to our work is to use so-called "leaky integration" neurons~\citep{jaeger2007optimization},
which also forces the neurons to change their state slowly, however without the structural constraint of SCRN.
It was evaluated on the same dataset as we use further (Penn Treebank) by~\cite{bengio2013advances}. Interestingly,
the results we observed in our experiments show much bigger gains over stronger baseline using our model, as will
be shown later.

\paragraph{Alternative Model Interpretation.}
If we consider the context units as additional hidden units (with no activation function),
we can see our model as a SRN with a constrained recurrent matrix $M$ on both hidden and context units:

\begin{equation}
\label{ref:alt}
M=\left[
\begin{array}{c|c}
R & P \\ \hline
0 &\alpha I_p
\end{array}\right],
 \end{equation}
 
where $I_p$ is the identity matrix and $M$ is a square matrix of size $m+p$, i.e., the sum of the number of hidden and context units.
This reformulation shows explicitly our structural modification of the Elman SRN~\citep{elman1990finding}: we constrain a diagonal 
block of the recurrent matrix to be equal to a reweighed identity, and keep an off-diagonal block equal to 0. For this reason,
we call our model \emph{Structurally Constrained Recurrent Network} (SCRN).

\paragraph{Adaptive Context Features.} 
Fixing the weight $\alpha$ to be constant in Eq.~(\ref{eq:s}) forces the hidden units to capture information on the same time scale.
On the other hand, if we allow this weight to be learned for each unit, we can potentially capture context from different time delays~\citep{pachitariu2013regularization}. 
More precisely, we denote by $Q$ the recurrent matrix of the contextual hidden layer, and we consider the following update rule for the state of the contextual hidden layer $s_t$:

\begin{equation}
s_t = (I - Q) B x_t + Q s_{t-1}, 
\label{eq:alt}
\end{equation}

where $Q$ is a diagonal matrix with diagonal elements in $(0,1)$. We suppose that
these diagonal elements are obtained by applying a sigmoid transformation to a parameter vector $\beta$, 
i.e., diag$(Q)=\sigma(\beta)$. This parametrization naturally forces the diagonal weights to stay strictly between 0 and 1.

We study in the following section in what situations does learning of the weights help.
Interestingly, we show that learning of the self-recurrent weights does not seem to be important, as long as one uses also the standard hidden layer in the model.

%% file: experiments.tex
\section{Experiments}
\label{sec:exp}

We evaluate our model on the language modeling task for two datasets.
The first dataset is the Penn Treebank Corpus,
which consists of 930K words in the training set.
The pre-processing of data and division to training, validation and test parts are the same as in~\citep{mikolov2011extensions}.
The state-of-the-art performance on this dataset has been achieved by~\citet{zaremba2014recurrent}, using combination of many
big, regularized LSTM recurrent neural network language models. The LSTM networks were first introduced to language modeling by~\citet{sundermeyer2012lstm}.

The second dataset, which is moderately sized, is called Text8. It is composed of a pre-processed
version of the first 100 million characters from Wikipedia dump. We did split it into training part (first 99M characters) and development set (last 1M characters)
that we use to report performance. After that, we constructed the vocabulary and replaced all words that occur less than 5 times by $<$UNK$>$ token. The resulting
vocabulary size is about 44K.
To simplify reproducibility of our results, we released both the SCRN code and the scripts which construct the datasets \footnote{The SCRN code can be downloaded at \url{http://github.com/facebook/SCRNNs}}.

In this section we compare the performance of our proposed model against standard SRNs, and LSTM RNNs which are
becoming the architecture of choice for modeling sequential data with long-term dependencies.

\subsection{Implementation Details.}

We used Torch library and implemented our proposed model following the graph given in Figure \ref{fig:model}-b. Note that following the alternative interpretation
of our model with the recurrent matrix defined in Eq. \ref{eq:alt}, our model could be simply implemented by modifying a standard SRN.
We fix $\alpha$ at 0.95 unless stated otherwise. 
The number of backpropagation through time (BPTT) steps is set
to 50 for our model and was chosen by parameter search on the validation set. For normal SRN, we use just 10 BPTT steps because the gradients vanish faster. 
We do a stochastic gradient descent after every 5 forward steps.
Our model is trained with a batch gradient descent of size 32, and a learning rate of 0.05. We divide the learning rate by 1.5 
after each training epoch when the validation error does not decrease.


\subsection{Results on Penn Treebank Corpus.}

We first report results on the Penn Treebank Corpus using both small and moderately sized models (with respect to the number of hidden units).
Table \ref{tab:small} shows that our structurally constrained recurrent network (SCRN) model
can achieve performance comparable with LSTM models on small datasets with relatively small numbers of parameters. It should be noted
that the LSTM models have significantly more parameters for the same size of hidden layer, making the comparison somewhat unfair - with the
input, forget and output gates, the LSTM has about 4x more parameters than SRN with the same size of hidden layer. 

Comparison to "leaky neurons" is also in favor of SCRN: \cite{bengio2013advances} report perplexity reduction from 136 (SRN) to 129 (SRN + leaky neurons),
while for the same dataset, we observed much bigger improvement, going from perplexity 129 (SRN) down to 115 (SCRN).

Table \ref{tab:small} also shows that SCRN outperforms the SRN architecture even with much less parameters.
This can be seen by comparing performance of SCRN with 40 hidden and 10 contextual units (test perplexity 127) versus SRN with 300 hidden units (perplexity 129).
This suggests that imposing a structure on the recurrent matrix allows the learning algorithm to capture additional information. To obtain further evidence that this additional
information is of a longer term character, we did further run experiments on the Text8 dataset that contains various topics, and thus the longer term information
affects the performance on this dataset much more.

\begin{table}[h!]
\begin{center}
\begin{tabular}{c|c c | c c}
Model & $\#$hidden & $\#$context & Validation Perplexity & Test Perplexity\\
\hline
Ngram &- &- & - & 141\\
Ngram + cache	&- &-&-&125		\\
\hline
SRN 	&	50	& - & 153	& 144\\
SRN		&	100	& - & 137 & 129 \\
SRN		&	300	& - & 133 & 129 \\
\hline
LSTM 	&	50	& - &	129 & 123\\
LSTM 	&	100	& - & \textbf{120} & \textbf{115} \\
LSTM 	&	300	& - & 123 & 119 \\
\hline
SCRN 	& 	40	&10&133 & 127	\\
SCRN 	&	90	&10&124 &	119\\
SCRN	&	100	&40& \textbf{120} &  \textbf{115} \\
SCRN	&	300	&40& \textbf{120} &  \textbf{115} 
\end{tabular}
\caption{Results on Penn Treebank Corpus: n-gram baseline, simple recurrent nets (SRN), long short term memory RNNs (LSTM) and structurally constrained recurrent nets (SCRN).
Note that LSTM models have 4x more parameters than SRNs for the same size of hidden layer.}
\label{tab:small}
\end{center}
\end{table}

\subsubsection{Learning Self-Recurrent Weights.}

We evaluate influence of learning the diagonal weights of the recurrent matrix
for the contextual layer. For the following experiments, we used a hierarchical soft-max with 100 frequency-based classes on the Penn Treebank Corpus to speedup the experiments.
In Table~\ref{tab:adap}, we show that when the size of the hidden layer is small,
learning the diagonal weights is crucial. This result confirms the findings in~\cite{pachitariu2013regularization}. However, 
as we increase the size of our model and use sufficient number of hidden units,
learning of the self-recurrent weights does not give any significant improvement. This
indicates that learning the weights of the contextual units allows these units to
be used as multi-scale representation of the history, i.e., some contextual units can specialize on the very recent
history (for example, for $\alpha$ close to $0$, the contextual units would be part of a simple bigram language model). 
With various learned self-recurrent weights, the model can be seen as a combination of cache and bigram models.
When the number of standard hidden units is enough to capture short term patterns,
learning the self-recurrent weights does not seem crucial anymore.

Keeping this observation in mind we fixed the diagonal weights when working with the Text8 corpus. 

\begin{table}[h!]
\begin{center}
\begin{tabular}{c | c c |c c}
Model & $\#$hidden & $\#$context & Fixed weights &	Adaptive weights \\
\hline
SCRN & 50 & 0 &	 156 &	156\\
SCRN & 25 &25 & 150 & \textbf{145}\\
SCRN & 0 & 50 & 344 & 157 \\
\hline
SCRN & 140 & 0 &	 140 &	140\\
SCRN & 100 & 40 & \textbf{127} & \textbf{127}	\\
SCRN & 0 & 140 &	 334 &	147	
\end{tabular}
\caption{Perplexity on the test set of Penn Treebank Corpus with and without learning the weights of the contextual features. Note that in these experiments we used a hierarchical soft-max.}
\label{tab:adap}
\end{center}
\end{table}

\subsection{Results on Text8.}

Our next experiment involves the Text8 corpus which is significantly larger than
the Penn Treebank. As this dataset contains various articles from Wikipedia, the
longer term information (such as current topic) plays bigger role than in the previous
experiments. This is illustrated by the gains when cache is added to the baseline
5-gram model: the perplexity drops from 309 to 229 (26\% reduction).

We report experiments with a range of model configurations,
with different number of hidden units. In Table~\ref{tab:text8-slow}, we show that increasing
the capacity of standard SRNs by adding the contextual features
results in better performance. For example, when we add 40 contextual units to SRN with 100 hidden units,
the perplexity drops from 245 to 189 (23\% reduction). Such model is also much better than SRN with
300 hidden units (perplexity 202).

\begin{table}[h!]
\begin{center}
\begin{tabular}{c | c | c c c c c}
Model & $\#$hidden & context = 0 & context = 10 & context = 20 & context = 40 & context = 80\\
\hline
SCRN & 100 &	 245 & 215 & 201	& 189 &  184\\
SCRN & 300 &	 202 & 182 & 172 	& 165 & 164	\\
SCRN & 500 &	184 & 177  & 166 	& 162 & \textbf{161} 		
\end{tabular}
\caption{Structurally constrained recurrent nets: perplexity for various sizes of the contextual layer, reported on the development set of Text8 dataset.}
\label{tab:text8-slow}
\end{center}
\end{table}

In Table \ref{tab:text8}, we see that when the number of
hidden units is small, our model is better than LSTM. Despite the LSTM model with 100 hidden units
being larger, the SCRN with 100 hidden and 80 contextual features achieves better performance.
On the other hand, as the size of the models increase,
we see that the best LSTM model is slightly better than the best SCRN (perplexity 156 versus 161).
As the perplexity gains for both LSTM and SCRN over SRN are much more significant than in the Penn Treebank
experiments, it seems likely that both models actually model the same kind of patterns in
language.

\begin{table}[h!]
\begin{center}
\begin{tabular}{c |c c | c}
Model & $\#$hidden & $\#$context & Perplexity on development set \\
\hline
SRN		&	100	& - & 245	\\
SRN		&	300	& - & 202  \\
SRN		&	500	& - & 184  \\
\hline
LSTM & 100& - &	193 \\				
LSTM & 300& - &	159 \\				
LSTM & 500& - &   	\textbf{156} \\
\hline		
SCRN		&	100	& 80 & 184\\
SCRN		&	300	& 80 & 164  \\
SCRN		&	500	& 80 & 161  \\
\end{tabular}
\caption{Comparison of various recurrent network architectures, evaluated on the development set of Text8.}
\label{tab:text8}
\end{center}
\end{table}

%% file: conclusion.tex
\section{Conclusion}

In this paper, we have shown that learning longer term patterns in real data using recurrent networks
is perfectly doable using standard stochastic gradient descent, just by introducing structural constraint
on the recurrent weight matrix. The model can then be interpreted as having quickly changing hidden layer
that focuses on short term patterns, and slowly updating context layer that retains longer term information.

Empirical comparison of SCRN to Long Short Term Memory (LSTM) recurrent network shows very similar behavior in two
language modeling tasks, with similar gains over simple recurrent network when all models are tuned for the best
accuracy. Moreover, SCRN shines in cases when the size of models is constrained, and with similar number of parameters
it often outperforms LSTM by a large margin. This can be especially useful in cases when the amount of training
data is practically unlimited, and even models with thousands of hidden neurons severely underfit the training
datasets.

We believe these findings will help researchers to better understand
the problem of learning longer term memory in sequential data. Our model greatly
simplifies analysis and implementation of recurrent networks that are capable of learning longer term patterns.
Further, we published the code that allows to easily reproduce experiments described in this paper.

At the same time, it should be noted that none of the above models is capable of learning truly
long term memory, which has a different nature. For example, if we would want to build a model
that can store arbitrarily long sequences of symbols and reproduce these later, it would become obvious
that this is not doable with models that have finite capacity. A possible solution is to use the recurrent
net as a controller of an external memory which has unlimited capacity. For example in \citep{joulin2015inferring}, a stack-based
memory is used for such task. However, a lot of research needs to be done in this direction before we will
develop models that can successfully learn to grow in complexity and size when solving increasingly more difficult tasks.